\pdfoutput=1

\documentclass[11pt]{article}

\usepackage[]{EMNLP2022}

\usepackage{times}
\usepackage{latexsym}

\usepackage[T1]{fontenc}

\usepackage[utf8]{inputenc}

\usepackage{microtype}

\usepackage{inconsolata}

\title{Challenges and Applications of Automated Extraction of Socio-political Events from Text (CASE 2022): Workshop and Shared Task Report}

\author{Ali Hürriyetoğlu \\
  KNAW Humanities Cluster DHLab \\
  \texttt{\small ali.hurriyetoglu@dh.huc.knaw.nl} \\\And
  Hristo Tanev \\
  European Commission \\
  \texttt{\small hristo.tanev@ec.europa.eu} \\\And
  Vanni Zavarella \\
  University of Cagliari \\
  \texttt{\small v.zavarella@unica.it} \\\AND
  Reyyan Yeniterzi \\
  Sabanci University \\
  \texttt{\small reyyan.yeniterzi@sabanciuniv.edu} \\\And
  Osman Mutlu \\
  Koc University \\
  \texttt{\small omutlu@ku.edu.tr} \\\And
  Erdem Yörük \\
  Koc University \\
  \texttt{\small eryoruk@ku.edu.tr}}

\begin{document}
\maketitle
\begin{abstract}
We provide a summary of the fifth edition of the CASE workshop that is held in the scope of EMNLP 2022. The workshop consists of regular papers, two keynotes, working papers of shared task participants, and task overview papers. This workshop has been bringing together all aspects of event information collection across technical and social science fields. In addition to the progress in depth, the submission and acceptance of multimodal approaches show the widening of this interdisciplinary research topic. 
\end{abstract}

\section{Introduction}

The workshop on Challenges and Applications of Automated Extraction of Socio-political Events from Text (CASE) has become a significant venue where all technical and social science aspects of event information collection can be discussed in its fifth edition.\footnote{\url{https://emw.ku.edu.tr/case-2022/}, accessed on November 14, 2022.} The 2022 Conference on Empirical Methods in Natural Language Processing (EMNLP 2022) hosts the edition this year between December 7 and 11 in Abu Dhabi .\footnote{\url{https://2022.emnlp.org/}, accessed on November 14, 2022.}

Socio-political event extraction (SPE) has long been a challenge for the natural language processing (NLP) community as it requires sophisticated methods in defining event ontologies, creating language resources, and developing algorithmic approaches~\cite{Pustejovsky+2003,Boros2018,Chen+2021}. Social and political scientists have been working to create socio-political event (SPE) databases such as ACLED, EMBERS, GDELT, ICEWS, MMAD, PHOENIX, POLDEM, SPEED, TERRIER, and UCDP following similar steps for decades. These projects and the new ones increasingly rely on machine learning (ML), deep learning (DL), and NLP methods to deal better with the vast amount and variety of data in this domain~\cite{hurriyetoglu-etal-2020-automated,hurriyetoglu-etal-2021-challenges}. Automation offers scholars not only the opportunity to improve existing practices but also to vastly expand the scope of data that can be collected and studied, thus potentially opening up new research frontiers within the field of SPEs, such as political violence and social movements. But automated approaches  suffer from major issues like bias, generalizability, class imbalance, training data limitations, and ethical issues that have the potential to affect the results of automated text processing systems and their use drastically~\cite{Leins+2020,Bhatia+2020,Chang+2019}. Moreover, the results of the automated systems for SPE information collection have neither been comparable to each other nor been of sufficient quality~\cite{Wang+16,Schrodt2020}. 
 
Setting a clear path toward addressing these challenges is our main focus. We are confident that the program we put together for this year's event after rigorous and thorough reviews, would bring us closer to that goal and beyond. 

We provide a summary of the accepted papers in the following Section, which is Section~\ref{sec:accepted-papers}. Next, the shared tasks that are organized in the scope of this workshop are described in Section~\ref{sec:tasks}. The keynote abstracts and invited talks are provided in sections~\ref{sec:keynotes} and~\ref{sec:invited}. Finally, Section~\ref{sec:conclusion} conclude this report with a brief summary and future outlook.

\section{Accepted papers}
\label{sec:accepted-papers}

This year, out of 12 submissions 8 were accepted by the program committee. A quick summary of these papers are provided below.

\begin{itemize}
    \item \citet{thapa-etal-2022-a-multimodal} releases a multimodal dataset that consists of
of 5,680 text-image pairs of tweets and a baseline for hate
speech detection in the context of Russia-Ukraine war.
    \item \citet{you-etal-2022-eventgraph} propose Event-Graph, a joint framework for event extraction,
which encodes events as graphs. They represent
event triggers and arguments as nodes in a
semantic graph. Event extraction therefore becomes
a graph parsing problem.
    \item \citet{desot-etal-2022-a-hybrid} suggests event and argument role detection as one task
in a hybrid event detection approach and a novel
method for automatic self-attention threshold
selection. 
    \item \citet{mehta-etal-2022-improving} cast socio-political conflict event extraction as a machine reading comprehension (MRC) task. In this approach, extraction of socio-political actors and targets from a sentence is framed as an extractive question answering problem conditioned on event type.
    \item \citet{ria-etal-2022-cross-modal} propose a method that utilizes existing annotated unimodal data to perform event detection in another data modality using a zero-shot setting. They focus on protest detection in text and images, and show that a pretrained vision-and-language alignment model (CLIP) can be leveraged towards this end.
    \item \citet{sticha-etal-2022-hybrid} release a comprehensive, consolidated, and cohesive assassination dataset that is prepared utilizing a robust ML framework that prioritizes understandability through visualizations and generalizability through the ability to implement different ML algorithms. 
    \item \citet{dai-etal-2022-political} train a model that can produce structured political event records at the sentence level. This approach is based on text-to-text sequence generation. They also describe a method for generating synthetic text and event record pairs that we use to fit a model.
    \item \citet{akdemir-etal-2022-zero-shot} approach the classification problem as an entailment problem and apply zero-shot ranking to socio-political texts. Documents that are ranked at the top can be considered positively classified documents and this reduces the close reading time for the information extraction process.
\end{itemize}

\section{Shared tasks}
\label{sec:tasks}

Three tasks were organized in the scope of CASE 2022. Each one of these tasks shed light on a different aspect of event information collection. These are zero-shot and detailed multilingual event information, evaluation of state-of-the-art systems in replicating manually curated event datasets, and event causality detection.

\subsection{Task 1: Extended Multilingual Protest Event Detection}

The extended multilingual protest news detection is the same shared task organized at CASE 2021~\cite{hurriyetoglu-etal-2021-multilingual}. This year we introduced additional data and languages at the evaluation stage.\footnote{\url{https://github.com/emerging-welfare/case-2022-multilingual-event}, accessed on November 15, 2022.} This year, the Task 1 focused on evaluating the zero-shot prediction performance of the state-of-the-art systems for Subtask 1, document classification. The training set is the same with CASE 2021 data that is in English, Portuguese, and Spanish. But the evaluation data consists of the union of CASE 2021 test data and new data in both available and new languages. The new languages are Mandarin, Urdu, and Turkish. Details of CASE 2022 Task 1 is reported by~\citet{hurriyetoglu-etal-2022-extended}.

\subsection{Task 2: Tracking COVID-19 protest events in the United States}

This task aims at automatically replicating manually created event datasets. The participants of Task 1 are invited to run the systems they develop to tackle Task 1 on a news and a Twitter archive. This is a similar setting with the edition performed last year in the scope of CASE 2021 and reported by ~\citet{giorgi-etal-2021-discovering}. This year's results \footnote{\url{https://github.com/zavavan/case2022_task2}, accessed on November 14, 2022.} are reported by~\citet{zavarella-etal-2022-covid19}.

\subsection{Task 3: Event Causality identification}

Causality is a core cognitive concept and appears in many natural language processing (NLP) works that aim to tackle inference and understanding. This task focuses on the study of event causality in news, and therefore, introduces the Causal News Corpus~\cite{tan-etal-2022-the-causal}. The Causal News Corpus consists of 3,559 event sentences from CASE 2021 data, extracted from protest event news, that have been annotated with sequence labels on whether it contains causal relations or not. Subsequently, causal sentences are also annotated with Cause, Effect, and Signal spans. The two subtasks (Sequence Classification and Span Detection) work on the Causal News Corpus, and accurate, automated solutions are proposed for the detection and extraction of causal events in news. The detailed report of the task is provided in \citet{tan-etal-2022-event}.~\footnote{\url{https://github.com/tanfiona/CausalNewsCorpus}, accessed on November 14, 2022.} 

\section{Keynotes}
\label{sec:keynotes}

Three scholars delivered two keynote speeches that are one on event extraction system development and one for error analysis of event information collection systems. The speakers were invited according to our tradition of having one keynote with technical and another one with social and political sciences, background. We provide abstracts of the keynote speeches as they are provided by the keynote speakers in the following subsections. Section~\ref{sec:nguyen} and Section~\ref{sec:althaus} are contributions of Prof. Thien Huu Nguyen~\footnote{\url{https://ix.cs.uoregon.edu/~thien/}, accessed on November 14, 2022.} and Prof. Scott Althaus\footnote{\url{https://ix.cs.uoregon.edu/~thien/}, accessed on November 14, 2022.} and Prof. J. Craig Jenkins\footnote{\url{https://sociology.osu.edu/people/jenkins.12}, accessed on November 15, 2022.} respectively.\footnote{The personal pronoun usege such as `I' and `we' in the following subsections indicate the keynote speakers and not the authors of this report.}

\subsection{Event Extraction in the Era of Large Language Models: Structure Induction and Multilingual Learning}
\label{sec:nguyen}

Events such as protests, disease outbreaks, and natural disasters are prevalent in text from different languages and domains. Event Extraction (EE) is an important task of Information Extraction that aims to identify events and their structures in unstructured text. The last decade has witnessed significant progress for EE research, featuring deep learning and large language models as the state-of-the-art technologies. However, a key issue of existing EE methods involves modeling input text sequentially to solve each EE tasks separately, thus limiting the abilities to encode long text and capture various types of dependencies to improve EE performance. In this talk, I will present some of our recent efforts to address this issue where text structures are explicitly learned to realize important objects and their interactions to facilitate the predictions for EE.

In addition, current EE research still mainly focuses on a few popular languages, e.g., English, Chinese, Arabic, and Spanish, leaving many other languages unexplored for EE. In this talk, I will also introduce our current research focus on developing evaluation benchmarks and models to extend EE systems to multiple new languages, i.e., multilingual and cross-lingual learning for EE. Finally, I will highlight some research challenges that can be studied in future work for EE.

\subsection{A total error approach to validating event data that is transparent, scalable, and practical to implement}
\label{sec:althaus}

There are at least two reasonable ways to make your way toward where you want to go: looking down to carefully place one foot in front of the other, and looking up to focus on where you hope to arrive. Looking up beats looking down if there’s a particular destination in mind, and for constructing valid event data that destination usually takes the form of high-quality human judgment. Yet many approaches to generating event data on protests and acts of political violence using fully-automated systems implicitly adopt a “looking down” approach by benchmarking validity as a series of incremental improvements over prior algorithmic efforts. And even those efforts that adopt a ``looking up'' approach often treat human-generated gold standard data as if it was prima facie valid, without ever testing or confirming the accuracy of this assumption. It stands to reason that if we want to automatically produce valid event data that approaches the validity of human judgment, then we also need to validate the human judgment tasks that provide the point of comparison. But because of obvious difficulties in implementing such a rigorous assessment within the time and budget constraints of typical research projects, this more rigorous double-validation approach is rarely attempted.

This presentation outlines a ``looking up'' approach for double-validating fully-automated event data developed by the Cline Center for Advanced Social Research at the University of Illinois Urbana-Champaign (USA), illustrates that approach with a test of the precision and recall for two widely-used event classification systems (the PETRARCH-2 coder used in Phoenix and TERRIER, as well as the BBN ACCENT coder used in W-ICEWS), and demonstrates the utility of the approach for developing fully-automated event data algorithms with levels of validity that approach the quality of human judgment. 

The first part of the talk reviews the Cline Center’s total error framework for identifying 19 types of error that can affect the validity of event data and addresses the challenge of applying a total error framework when authoritative ground truth about the actual distribution of relevant events is lacking~\cite{althaus-etal-2022-a-total}. We argue that carefully constructed gold standard datasets can effectively benchmark validity problems even in the absence of ground truth data about event populations. We propose that a strong validity assessment for event data should, at a minimum, possess three characteristics. First, there should be a standard describing ideal data; a gold standard that, in the best case, takes the form of ground truth. Second, there should be a direct “apples to apples” comparison of outputs from competing methods given identical input. Third, the test should use appropriate metrics for measuring agreement between the gold standard and data produced by competing approaches.

The second part of the talk presents the results of a validation exercise meeting all three criteria that is applied to two algorithmic event data pipelines: the Python Engine for Text Resolution and Related Coding Hierarchy (PETRARCH-2) and the BBN ACCENT event coder. It then reviews a recent Cline Center project that has built a fully-automated event coder which produces dramatic improvements in validity over both PETRARCH-2 and BBN ACCENT by leveraging the total error framework and a reliance on the double-validation approach using high-quality gold standard benchmark datasets.

\section{Invited talks}
\label{sec:invited}

Papers that are accepted to be published in the Findings of EMNLP 2022 and related to our workshop theme were invited to be presented during our workshop. The authors of the following papers were invited for presenting their papers:

\begin{itemize}
    \item ~\citet{Jiao+2022} define the task open-vocabulary argument role prediction. The goal of this task is to infer a set of argument roles for a given event type. They propose a novel unsupervised framework, \textsc{RolePred} for this task and release a new human-annotated event extraction dataset including 139 customized argument roles with rich semantics. 
    \item ~\citet{Faghihi+2022} presents CrisisLTLSum, the largest dataset of local crisis event timelines about wildfires, local fires, traffic, and storms available to date. CrisisLTLSum was built using a semi-automated cluster-then-refine approach to collect data from the public Twitter stream. 
    \item ~\citet{Ding+2022} design a neural model that they refer to as the Explicit Role Interaction Network (ERIN) which allows for dynamically capturing the correlations between different argument roles within an event. 
    \item ~\citet{Gao+2022} present Mask-then-Fill, a flexible and effective data augmentation framework for event extraction. This approach allows for more flexible manipulation of text and thus can generate more diverse data while keeping the original event structure unchanged. 
\end{itemize}

\section{Conclusion}
\label{sec:conclusion}

The CASE workshop series has been contributing to both technical advancement in terms of shared task organization and being a venue for scholars working at the intersection of social sciences and event extraction. This role become more significant as these series are known to a wider community. Following steps of this series should serve the community by preserving its inderdisciplinary setting, welcoming new methodologies, and promoting responsible development and utilization of the results of this scholarship.

\bibliography{anthology,custom}
\bibliographystyle{acl_natbib}




\end{document}